\documentclass{article}

 \usepackage[preprint]{neurips_2025}

\usepackage{enumitem}
\usepackage[utf8]{inputenc} 
\usepackage[T1]{fontenc}    
\usepackage{url}            
\usepackage{booktabs}       
\usepackage{amsfonts}       
\usepackage{nicefrac}       
\usepackage{microtype}      
\usepackage{xcolor}         
\usepackage{makecell} 
\usepackage{adjustbox}
\usepackage{pifont}
\usepackage{bbding}
\usepackage{tabulary,multirow,xspace}
\usepackage{fixmath,mathtools,nicefrac,mmstyle}
\usepackage{subcaption}
\usepackage{caption}
\usepackage{float}
\usepackage{wrapfig} 
\usepackage[misc]{ifsym} 
\usepackage{colortbl}
\definecolor{mygray1}{gray}{.95}
\definecolor{mygray2}{gray}{.9}
\definecolor{mygray3}{gray}{.95}
\usepackage{pifont}
\newcommand{\cmark}{\ding{51}}%

\newlength\savewidth
\newcolumntype{x}[1]{>{\centering\arraybackslash}p{#1pt}}

\newcommand{\app}{\raise.17ex\hbox{$\scriptstyle\sim$}}

\definecolor{linkcolor}{RGB}{255,0,0}
\definecolor{urlcolor}{RGB}{255,105,180}
\definecolor{citecolor}{RGB}{66,168,235}
\usepackage[pagebackref=true,breaklinks=true,letterpaper=true,colorlinks,bookmarks=false,citecolor=blue,linkcolor=blue]{hyperref}

\newcommand \footnoteONLYtext[1]
{
	\let \mybackup \thefootnote
	\let \thefootnote \relax
	\footnotetext{#1}
	\let \thefootnote \mybackup
	\let \mybackup \imareallyundefinedcommand
}

\title{Mixed-R1: Unified Reward Perspective For Reasoning Capability in Multimodal Large Language Models}
\author{Shilin Xu$^{1,2}$,Yanwei Li$^{1}$, Rui Yang$^{1}$, Tao Zhang$^{1}$, Yueyi Sun$^{2}$,Wei Chow$^{1}$,\\
\textbf{Linfeng Li$^{1}$, Hang Song$^{1}$, Qi Xu$^{1}$, Yunhai Tong$^{2}$, Xiangtai Li$^{1}$,  Hao Fei$^{3}$}\\
  {$^{1}$ByteDance} \quad
  {$^{2}$Peking University} \quad
  {$^{3}$National University of Singapore}
}

\begin{document}

\maketitle

\begin{abstract}
Recent works on large language models (LLMs) have successfully demonstrated the emergence of reasoning capabilities via reinforcement learning (RL).
Although recent efforts leverage group relative policy optimization (GRPO) for MLLMs post-training, they constantly explore one specific aspect, such as grounding tasks, math problems, or chart analysis. 
There are no works that can leverage multi-source MLLM tasks for stable reinforcement learning.
In this work, we present a unified perspective to solve this problem.
We present Mixed-R1, a unified yet straightforward framework that contains a mixed reward function design (Mixed-Reward) and a mixed post-training dataset (Mixed-45K).
We first design a data engine to select high-quality examples to build the Mixed-45K post-training dataset.
Then, we present a Mixed-Reward design, which contains various reward functions for various MLLM tasks.
In particular, it has four different reward functions: matching reward for binary answer or multiple-choice problems, chart reward for chart-aware datasets, IoU reward for grounding problems, and open-ended reward for long-form text responses such as caption datasets.
To handle the various long-form text content, we propose a new open-ended reward named Bidirectional Max-Average Similarity (BMAS) by leveraging tokenizer embedding matching between the generated response and the ground truth.
Extensive experiments show the effectiveness of our proposed method on various MLLMs, including Qwen2.5-VL and Intern-VL on various sizes.
Our dataset and model are available at \url{https://github.com/xushilin1/mixed-r1}.
\end{abstract}
\section{Introduction}
\label{sec:intro}
%

Recent advances in multimodal large language models~\cite{llava1.5, qwen2.5vl, internvl3, deepseek-vl, intern2.5, mllm-selector,vita,long-vita,cantor, yuan2025sa2va, zhang2025pixel, zhang2024omg,han2024free,evem,Jie2024MemorySpaceVP, zhou2025they} have facilitated the development of various MLLM benchmarks~\cite{ocrbench, mmstar,mmbench, mmmu, ai2d, mmvet,hallusionbench,seedbench,zhu2023genimage,videomme,mvbench,mluv,han2024free,videorag, fei2025path} via scaled pre-training, supervised fine-tuning, and post-training on massive multimodal datasets.
Despite effectiveness and stronger performance, several essential features, such as visual reasoning for more complex questions, are missing.
Meanwhile, recently, several large language models (LLMs), represented by OpenAI GPT-4o~\cite{gpt4o}, Claude 3.5, and DeepSeek-R1~\cite{r1}, have achieved groundbreaking progress in complex reasoning tasks, reaching human-expert levels in logical reasoning and mathematical problem-solving within textual contexts.
With step-by-step analysis and Chain-of-Thought (CoT) testing, these models ultimately obtain the correct solutions in mathematical and logical reasoning tasks.

Thus, recent MLLM works~\cite{mm-eureka,vlm-r1,videochart-r1,zhang2024enhancing, vision-r1,r1-onevision,r1v} also extend these breakthroughs from LLMs into vision-language tasks. 
Most works build on the DeepSeek-R1~\cite{r1} training pipeline, where they apply group relative policy optimization (GRPO) training.
For example, several works~\cite{r1v,vlm-r1,vision-r1} explore complex geometric relationships in images and conduct effective reasoning and proof with large-scale GRPO-like post-training, with high-quality math datasets.
Meanwhile, several works aim to solve the visual grounding~\cite{refcocog} tasks by introducing an explicit IoU reward or mAP (mean average precision) reward, where the distance serves as a reward signal to optimize the model's localization ability.
However, these works mainly focus on one specific topic in MLLM benchmarks, such as referring or math problems. 
For recent state-of-the-art MLLM foundation models, such as Qwen2.5-VL~\cite{qwen2.5vl} and InternVL 2.5~\cite{intern2.5}, they always obtain good performance on multiple general benchmarks, with a larger scope than those works for post-training.
For example, Qwen2.5-VL can be tested on over 30 benchmarks in various scenes.
As a result, there is a natural gap between task-specific RL post-training and general tasks in current MLLMs.

\begin{figure}[t]
    \centering
    \includegraphics[width=1.0\linewidth]{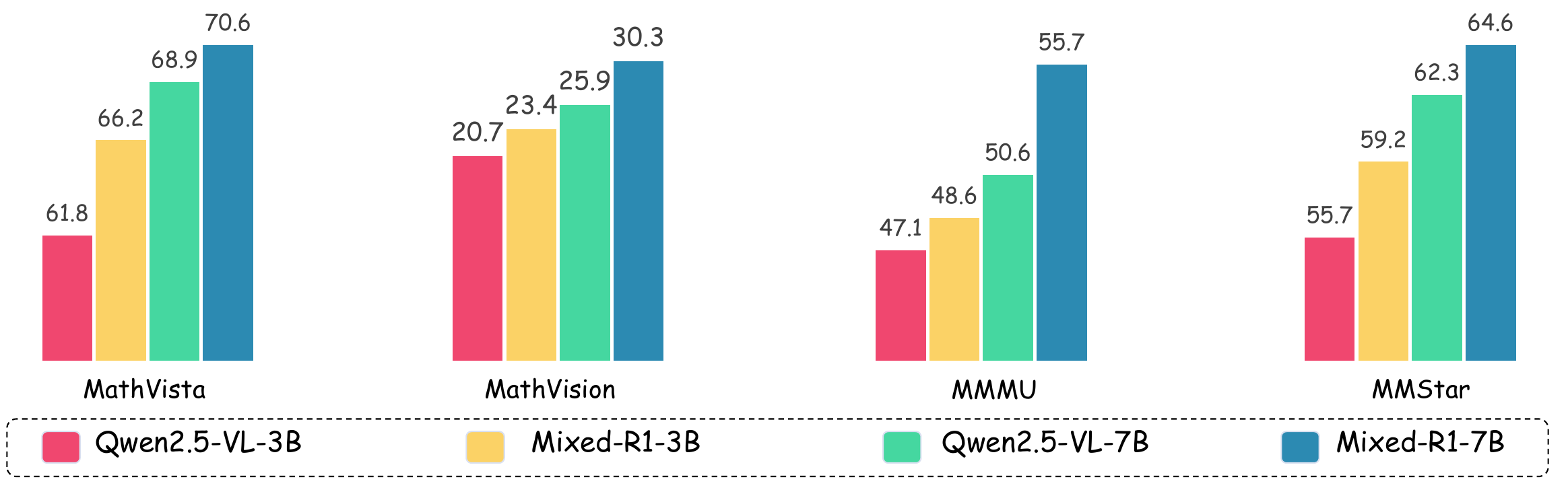}
    \caption{\textbf{Results of Our Mixed-R1}. Our post-training method improves Qwen2.5-VL on four different challenging benchmarks by large margins, on both the 3B model and the 7B model.}
    \label{fig:teaser}
\end{figure}

To solve this problem, we provide a more unified perspective with the reward function design and various post-training sample collections. 
Based on the DeepSeek-R1~\cite{r1} framework, we present Mixed-R1, the post-training framework to improve current SOTA MLLMs on multiple benchmarks.
According to the answer representation, we first revisit the current MLLM training set by dividing the existing training datasets into five categories: yes or no questions, multiple choice questions, chart and document-aware questions, referring and grounding questions, and open-ended questions.
With these five data types, we first collect a high-quality post-training dataset, Mixed-45K. 
It contains 45K high-quality examples that are sampled from existing open-sourced datasets.
In particular, we adopt Qwen2.5-VL-7B to filter examples to avoid either too simple or too hard questions.

Then, to perform the GRPO~\cite{r1} training on the mixed dataset, we propose a mixed reward function design to address various forms of data, particularly for open-ended datasets.
We divide the five data types into four different reward formats. 
In particular, according to different question types, the reward functions can be different. (See Fig.~\ref{fig:method})
We adopt a matching reward function according to the matching result for yes or no questions and multiple-choice questions.

For the chart-like dataset, where the answer is usually a number, we verify the accuracy of the results numerically.
We adopt an IoU reward to train the model for referring and grounding questions, where the difference between the prediction box and the ground truth box serves as the reward signal.
For open-ended responses, such as captions, we design a new BMAS reward function, stabilize the training, and improve performance on image understanding tasks.
In particular, rather than introducing extra LLM models as judges, we directly use the LLM model itself to obtain the embedding tokens for both outputs and ground truth. 
Then, we calculate the pairwise similarity between the generated text embedding and the ground truth text embedding.
Our key insight is to compute a bidirectional maximum similarity aggregation score by averaging the maximum similarity for each vector in both directions and summing the two scores. 
Using maximum similarity can avoid the ambiguities that are caused by different words with similar semantic meanings.
And then the final score serves as a direct reward signal.

Extensive experiments show the effectiveness of such a design on multiple benchmarks. 
We also carry out extensive ablation studies on the Mixed-45K dataset, the effectiveness of each reward, and the BMAS design.
As shown in Fig.~\ref{fig:teaser},  Mixed-R1 improves 2\%-5\% performance compared with Qwen2.5-VL models.

Our main contributions can be summarized as follows:
\begin{itemize}
\item We present Mixed-R1, a unified RL post-training framework to boost SOTA MLLMs via R1-like post-training, and it handles various data types simultaneously using a mixed reward design.

\item We propose a novel BMAS (Bidirectional Max-Average Similarity) based reward design to handle open-ended data like a caption dataset, without the need for any additional language model.

\item We provide a simple yet effective dataset, Mixed-45K, which serves as a post-training dataset for MLLMs.

\item Experiments show the Mixed-R1 has significant improvements on various complex reasoning benchmarks.
\end{itemize}
\section{Related Work}
\label{sec:related_work}
\noindent
\textbf{Multimodal Large Language Models.} Recent works~\cite{llava,llava1.5,qwenvl,qwen2.5vl,minicpm,luo2024feast,deepseek-vl,gemini,gpt4,qwen2.5-1m,luo2023cheap} have demonstrated remarkable capabilities in visual comprehension. 
These models excel at integrating and processing diverse data modalities, such as text, images, and videos, primarily attributed to large-scale pretraining and high-quality visual instruction tuning datasets.
In particular, open-sourced MLLMs adopt a LLaVA-like architecture: a visual encoder, an MLP projector, and a large language model (LLM). 
The visual encoder~\cite{clip, siglip, navit} extracts visual tokens from input images, while the MLP maps visual tokens into language tokens. 
The LLM~\cite{qwen, llama, vicuna, deepseekv3, intern2.5} then generates text responses by leveraging the visual tokens and the language input tokens.
Recent SOTA MLLMs~\cite{internvl3,llava-ov, qwen2.5vl, qwenvl, intern2.5} introduce a strong perception ability for high-resolution inputs.
Despite its effectiveness, there are still limitations on reasoning-aware MLLM tasks, including math tasks~\cite{mathvision,mathvista}, reasoning grounding tasks~\cite{refcocog}, and spatial understanding tasks~\cite{r1v}.
Recent works~\cite{r1v, videochart-r1} partially solve these problems for several specific tasks. 
However, no work can boost existing MLLMs in all directions.
Our work fills this gap by designing a unified reward function under the DeepSeek-R1 framework.
%
%

\noindent
\textbf{RL-based Post-training.}
The refinement of large language models (LLMs) has increasingly relied on reinforcement learning (RL) as a critical framework for aligning these systems with human preferences and task-specific goals. 
Works such as RLHF~\cite{rlhf}, PPO~\cite{ppo}, and DPO~\cite {dpo} have demonstrated remarkable success in enhancing the safety, coherence, and instruction-following capabilities of LLMs and MLLMs.
More recently, rule-based RL approaches, exemplified by GRPO, have shown significant promise for scaling RL applications across broader domains. Following this trend, MLLMs~\cite{r1v,vlm-r1,vision-r1,mm-eureka,r1-onevision} have rapidly adopted similar technologies, aiming to enhance their reasoning capabilities. 
However, these works are only simple attempts at rule-based GRPO, typically involving a simple dataset and a single reward function.
For example, R1-V~\cite{r1v} utilizes the GRPO~\cite{grpo} method and trains a model that targets object-counting tasks, and remarkably enables a 3B model to outperform a 72B model. 
VLM-R1 is one of the earliest works to successfully apply the GRPO~\cite{grpo} algorithm to multimodal models; however, this work primarily focuses on addressing detection issues.
Moreover, these methods cannot handle open-ended questions, such as those that require a reward for a caption. 
VideoChat-R1~\cite{videochart-r1} suggests using an LLM as a ``judge'' to assign a reward score.
They use the Qwen2.5-72B~\cite{qwen} to identify events in the description and assess if the model's predicted description encompasses the events in an accurate description. 
However, this requires an extra model, which costs more during training.
We propose a tokenizer-based approach to avoid such costs and stabilize the post-training process.
%
%

\section{Methodology}
\label{sec:method}

\subsection{Mixed-45K Dataset}
\label{sec:mixed_r1_45k}

In this section, we introduce the process of collecting and filtering training data to generate the Mixed-45K dataset. 
We present a simple data filter pipeline to generate high-quality data from open-sourced datasets.
In addition to data that can be directly judged using the filter rules, we also collect some open-ended data. 
We describe the types and distribution of Mixed-45K in the end.

\begin{figure}
    \centering
    \includegraphics[width=1.0\linewidth]{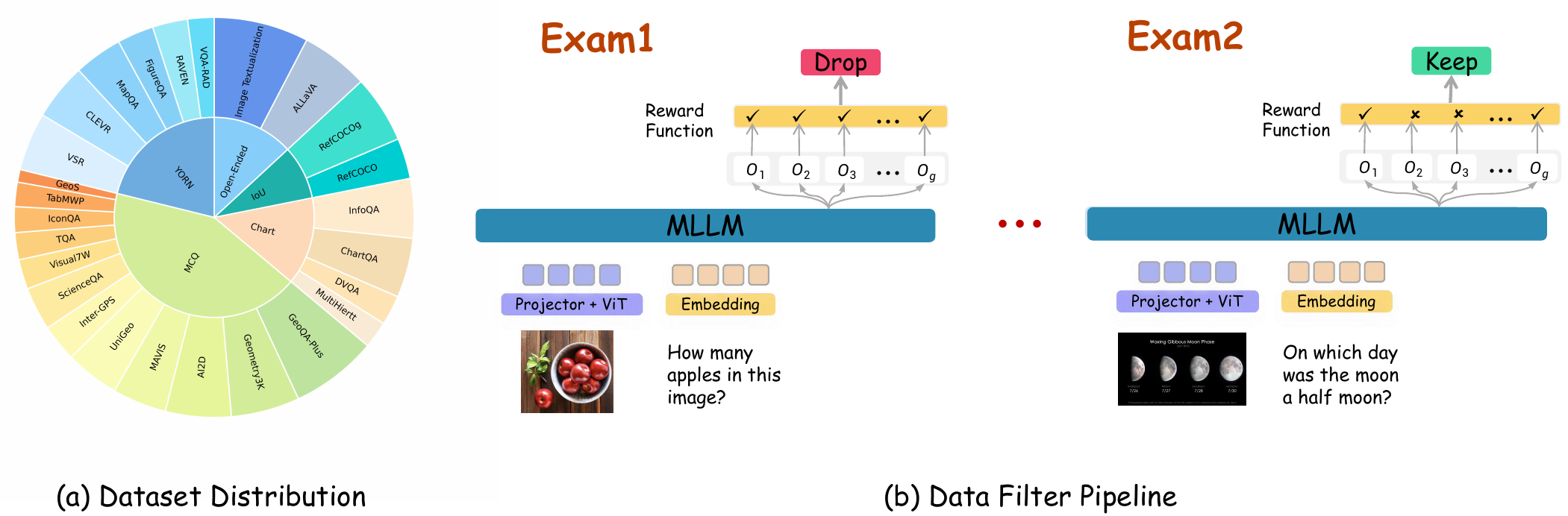}
    \caption{The distribution of the training dataset (a) and dataset filter pipeline (b).}
    \label{fig:stat}
\end{figure}

\textbf{Dataset Source and Dataset Filter.} 
We first introduce how to filter out the final training data. 
Firstly, we assume that all data conforms to the training format. 
We are only filtering out data that is more suitable for training in this process. 
Due to the calculation method of advantage in GPRO (Equ.~\ref{equ:adv}), the phenomenon of advantage disappearing occurs when the rewards of all responses are the same. 
To avoid this issue, we test the training dataset using Qwen2.5-VL~\cite{qwen2.5vl} in advance. 
Specifically, we set the temperature parameter to 1.0 and utilize Qwen2.5-VL-7B~\cite{qwen2.5vl} to generate $g$ responses for each question. By default, the value of $g$ is set to 8. If the rewards for each reply are identical, we will remove the corresponding data. 
This process involves deleting data that is either too simple or too hard.  And the Fig.~\ref{fig:stat} shows the final distribution of our Mixed-45K and the data filter pipeline.

\textbf{Dataset Types and Distribution.}
In summary, we have amassed a dataset of 45K samples:
\begin{itemize}[leftmargin=20pt]
    \item \textbf{YORN Dataset}: The YORN dataset requires the model to generate binary responses of ``yes'' or ``no''. Due to the relative simplicity of these data, they serve as a suitable starting point for GRPO training. We extracted approximately 9K samples from existing datasets such as MapQA and CLEVR~\cite{mapqa,clevr,figureqa,vsr}.

    \item \textbf{MCQ Dataset} The MCQ dataset consists of multiple-choice questions and is the largest category, containing around 20K samples. It includes a large portion of mathematical reasoning datasets~\cite{geoqa,unigeo} and general scientific reasoning datasets~\cite{ai2d,scienceqa}. This type of dataset evaluates a model's ability to understand and solve mathematical problems, as well as its capacity to follow logical reasoning. 
    
    \item \textbf{Chart-like Dataset}: These tasks challenge models to interpret complex, real-world visual and structured data representations, including charts, tables, and documents. These tasks require models not only to recognize and extract relevant information from diverse formats but also to reason about the underlying relationships and generate coherent, contextually appropriate responses. We collect such data samples from multiple data sources~\cite{chartqa,docvqa,infoqa,multihiertt}.

    \item \textbf{IoU Dataset}: These tasks assess a model's ability to ground natural language expressions within visual contexts. Specifically, models must identify and localize the specific object or region referred to in a sentence or phrase. This involves linguistic understanding and visual reasoning, often requiring the resolution of ambiguity, coreference, and contextual dependencies. We include datasets that emphasize compositional and cross-modal alignment~\cite{refcocog}
    
    \item \textbf{Open-Ended Dataset}: This task requires the model to observe and describe objects in the image, and a good caption also needs to infer the relationships between objects. For this task, we collect this dataset from ~\cite{allava}.
\end{itemize}

\subsection{Mixed-R1: Unified Reward Design}
\label{sub_sec:mixed_r1}

\noindent
\textbf{Preliminary of Group Relative Policy Optimization.}
Group Relative Policy Optimization (GRPO)~\cite{grpo} is a reinforcement learning (RL) technique that extends traditional policy optimization methods by incorporating relative comparisons between groups of policies.
By leveraging the rule-based reward, GRPO eliminates the commonly used critic model in other reinforcement learning methods, such as PPO~\cite{ppo}.
For each question $q$, GRPO first generates $G$ candidate responses $\{o_1, o_2, \dots , o_G \}$ from the old policy model $\pi_{\theta_{old}}$. For each candidate response $o$, GRPO obtains a score using a rule-based strategy, and then optimizes the policy model $\pi_{\theta}$ by maximizing the following objective:

\begin{align}
\mathcal{J}_{GRPO}(\theta) &= \mathbb{E}[q \sim P(Q), \{o_i\}_{i = 1}^{G} \sim \pi_{\theta_{old}}(O|q)] \nonumber \\
&\quad \frac{1}{G} \sum_{i = 1}^{G} \left( \min \left( \frac{\pi_{\theta}(o_i|q)}{\pi_{\theta_{old}}(o_i|q)} A_i, \mathrm{clip} \left( \frac{\pi_{\theta}(o_i|q)}{\pi_{\theta_{old}}(o_i|q)}, 1 - \varepsilon, 1 + \varepsilon \right) A_i \right) - \beta \mathbb{D}_{KL} \left( \pi_{\theta} || \pi_{ref} \right) \right),
\end{align}
where $\varepsilon$ and $\beta$ are hyper-parameters, and $A_i$ represents the advantage of the candidate response $o_i$ relative to other candidate responses.
And it is computed using a group of rewards $\{r_1, r_2,\dots, r_G\}$ corresponding to the outputs within each group:

\begin{align}
\label{equ:adv}
A_i=\frac{r_i - \text{mean}(\{r_1,r_2,\dots,r_G\})}{\text{std}(\{r_i,r_2,\dots,r_G\})}
\end{align}
For more detailed information, please refer to ~\cite{grpo, r1}.

\begin{figure}
    \centering
    \vspace{-5mm}
    \includegraphics[width=1.0\linewidth]{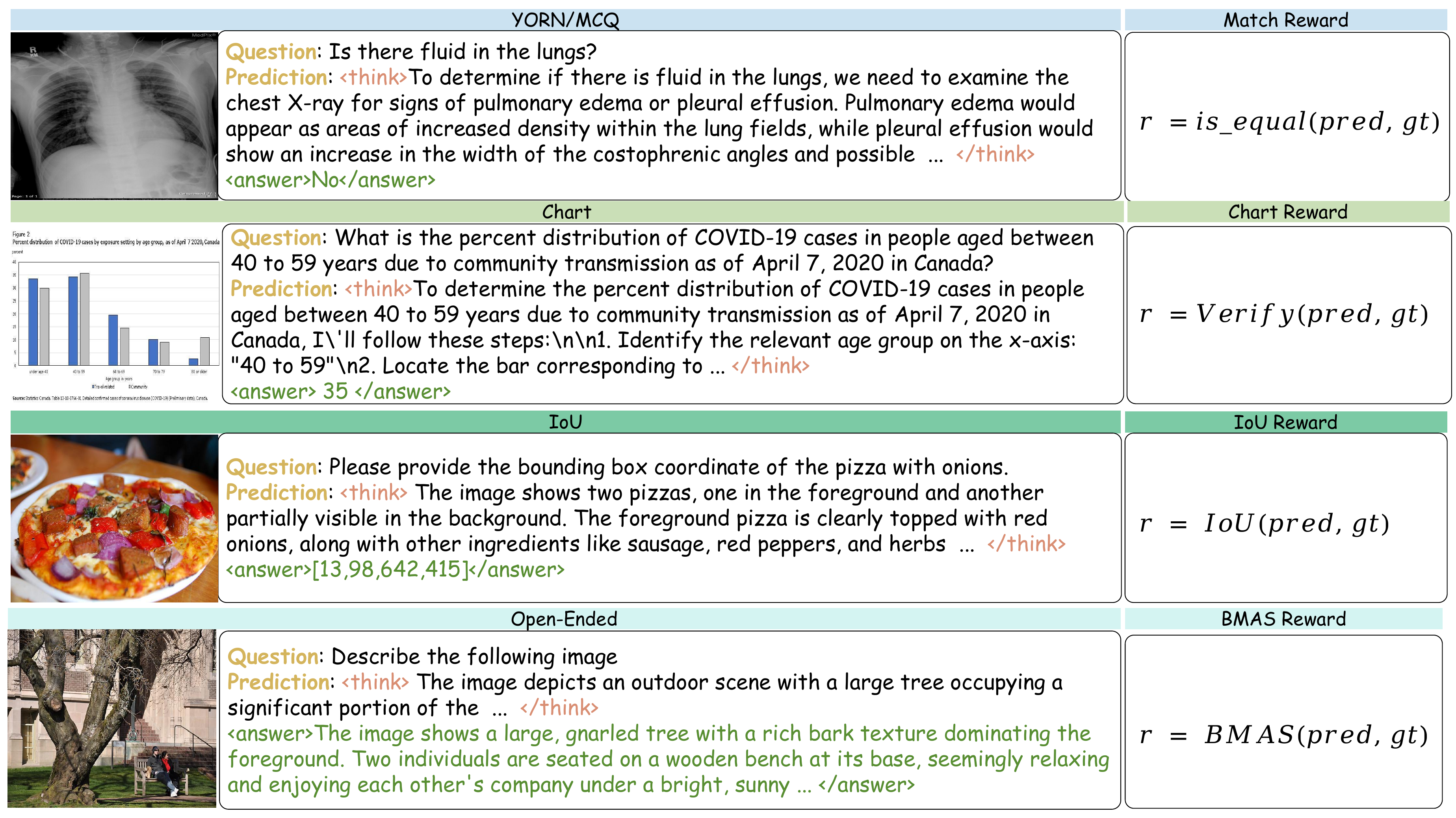}
    \caption{\textbf{Mixed-45K data types and reward examples} used in our Mixed-45K. }
    \label{fig:method}
\end{figure}

\noindent
\textbf{Unified Reward Function Design Driven by Data Types.} 
This work introduces the mixed reward training strategy, which includes multiple forms of training samples. Based on the different task types, we categorize the specific reward functions into the following groups (See the left examples in Fig.~\ref{fig:method}):

\begin{itemize}[leftmargin=20pt]
    \item \textbf{Matching Reward} means that the prediction can be directly compared to the ground truth to determine whether they are exactly the same. This type of reward can be applied to both the YORN and MCQ datasets. 
    \item \textbf{Chart Reward} is commonly used to evaluate datasets of charts, tables, or documents. Usually, we need to convert the prediction into floating-point numbers and verify the accuracy of the results.
    \item \textbf{IoU Reward} typically involves object detection datasets, which calculates the IoU of the predicted bounding box and the ground truth bounding box.
    \item \textbf{BMAS Reward} typically involves datasets that require a long-form text response. The specific calculation process can refer to the Equ~\ref{equ:tokenizer-max}
\end{itemize}

For matching reward, we check whether the response matches the ground truth for the YORN and MCQ data. We denote this reward as $r_{\text{matching}}$.
For the chart dataset, the model generally outputs numerical-type responses. We first convert the response into a floating-point number and then compute the error between the response and the ground truth value. 
If the error is less than 1e-2, the answer is considered correct. We denote this reward as  $r_{\text{chart}}$
For the above three types of questions, the value of the reward function is binary: the reward is 1 if the answer is correct, and 0 otherwise.

For the IoU reward, we first extract the bounding box of the object from the response and then calculate the IoU score with the ground truth bounding box. 
%
%
%
We use the IoU value between the prediction and the ground truth as the reward score.
We denote this reward as $r_{\text{IoU}}$.

\noindent
\textbf{BMAS Reward.}
We present an effective method for calculating rewards for open-ended questions, such as captions and open-ended answers (VQA).
In particular, we leverage the policy model to generate the language embedding and use the BMAS as the reward function.
%
Previous works~\cite{videochart-r1} introduce a large language model (LLM) as a ``judge'' to provide a reward score for each response. 
While effective, this strategy introduces an additional large model into the pipeline, which inevitably leads to increased computational overhead and limits scalability.

Our method addresses this limitation by enabling reward calculation using only the policy model $\pi_\theta$ itself, eliminating the need for an external LLM. 
Specifically, for each response $o$,  we first obtain the input embeddings $F_o \in \mathbb{R}^{N\times D}$ by mapping the original text into the input embedding space of the policy model, where $N$ is the number of response tokens and $D$ is the embedding dimension. 
Similarly, we obtain the corresponding input embeddings $F_g\in\mathbb{R}^{M\times D}$ for the ground truth answer, where $M$ denotes the number of tokens in the ground truth. 

To compute the reward, we measure the pairwise similarity $Sim\in \mathbb{R}^{N\times M}$ between the response and the ground truth based on their input embeddings. 
Each element $Sim_{i,j}$ represents the cosine similarity between the $i$-th token embedding in the response and the $j$-th token embedding in the ground truth:
\begin{align}
    Sim_{i,j}=\cos(F_o^i, F_g^j)
    \label{eq:cos}
\end{align}

We then aggregate these similarities to produce a final reward score. We compute the average of the maximum similarity values across each row and column:
\begin{align}
    \label{equ:tokenizer-max}
    r_{BMAS}(o,g) = \frac{1}{2} \left( \frac{1}{N} \sum_{i=1}^{N} \max_{j} Sim_{i,j} + \frac{1}{M} \sum_{j=1}^{M} \max_{i} Sim_{i,j} \right), 
\end{align}
where we find that simple maximum matching can work well, we denote this reward as $r_{BMAS}$

\noindent
\textbf{Format reward}
When answering all the aforementioned questions, the model uses a standard format reward as a starting point.
Specifically, we add a fixed prompt after each question. 
We incentivize outputs that provide a reasoning trace within the tags <think> ... </think> and a succinct final answer within the tag <answer> ... </answer>.
We apply a format reward, $r_{\text{format}}$, to each model response, assigning a reward value of 1 if the response conforms to the specified format and 0 otherwise.
\begin{equation}
    r_{\text{format}} = 
        \begin{cases}
        0, & \text{if output matches format}, \\
        1, & \text{if output doesn't match format}.
        \end{cases}
\end{equation}

\noindent
\textbf{Discussion.} During the development of Mixed-R1, we also tried extra judges like extra LLM or other LM, such as Bert~\cite{bert}. However, we find that all of them leave performance drops compared with baseline models. We argue that this may be due to semantic space inconsistency between the MLLM text tokenizer's output and the extra model's language embeddings. In addition, we also find that adopting bipartite graph matching for Equ .~\ref {eq:cos} also results in bad performances. This is because bipartite graph matching itself will forcefully assign each word embedding to the most matched words, leading to false positives and unstable training. More results can be found in Sec.~\ref{sec:exp_ablation}.

\noindent
\textbf{Final Reward Function.} We refer to the reward computed across all aforementioned data types. 
Then we obtain the final reward $r_{final}=r_\text{matching} + r_\text{chart} + r_\text{IoU} + r_\text{BMAS} + \lambda r_\text{format}$, where $\lambda$ is a hyperparameter that balances the contribution between format and answer. 
Unless otherwise specified, we set $\lambda=0.5$ by default for all experiments.

\section{Experiments}
\label{sec:exp}

\noindent
\textbf{Evaluation Benchmarks.} To comprehensively evaluate the model's performance, we conduct tests on multiple benchmarks based on lmms-eval~\cite{lmms-eval}. 
These benchmarks include MathVision~\cite{mathvision}, MathVista~\cite{mathvista}, MMBench~\cite{mmbench}, MMMU~\cite{mmmu}, MMStar~\cite{mmstar}, AI2D~\cite{ai2d}. 
These benchmarks collectively cover a broad spectrum of vision-language tasks, including mathematical reasoning and text recognition in images (OCR), enabling us to evaluate the model’s adaptability and performance across various application scenarios. 
The MathVision~\cite{mathvision} and MathVista~\cite{mathvision} are also curated collections of challenging mathematical problems that incorporate both textual and visual elements. 
%
MMStar~\cite{mmstar} emphasizes the model’s capability to interpret detailed visual content and answer questions requiring precise fine-grained analysis. 
Both AI2D~\cite{ai2d} and MMMU~\cite{mmmu} provide scientific or multi-disciplinary testing.
We report the average scores of these benchmarks, with a focus on the overall performance of the model on these benchmarks.

\noindent
\textbf{Implementation Details.} We mainly adopt the instruction-based Qwen2.5VL-7B~\cite{qwen2.5vl} as our baseline. 
Due to this series of models' powerful vision and text understanding abilities and instruction-following abilities, we can directly train them on the model without any cold start.
Thus, these features make it better to verify our mixed reward function design.
For the training hyperparameters, we set the training batch size to 64, with each sample generating eight responses. 
The temperature for model generation is set to 1 and 0.04 for the KL divergence.  For all the model training, we use a learning rate of 3e-6.
In addition, we also verify our method using the InternVL-2.5 4B baseline.

\subsection{Main Results}
\label{sec:exp_main_results}

\noindent
\textbf{Results on Various MLLMs.} In Tab.~\ref{tab:main_table}, we report our method on strong Qwen2.5-VL baselines, including the 3B model and the 7B model. We also report the SFT baseline and naive thinking inference for reference. 
Our method outperforms SFT baselines by 1.2\% for both 3B and 7B models, indicating the effectiveness of mixed RL reward.
Our method also performs better than the original baseline and the original baseline with thinking by around 2\%-3\% for both 3B and 7B models, on average.
In particular, our method works well for different types of data, indicating the unified improvement effects.

\begin{table}[t]
    \centering
    \setlength\tabcolsep{2pt}
    \adjustbox{max width=0.90\linewidth}{
    \begin{tabular}{l|ccccccccc}
    \toprule
    Method &  Thinking  & Avg. & MathVista & MathVision & MMBench & MMMU  & MMStar & AI2D \\
    \midrule
    \multirow{2}{*}{Qwen2.5-VL-3B} & - & 57.0 & 61.8 & 20.7 & 78.3 & 47.1 & 55.7 & 78.8 \\
    & \cmark & 53.2 & 55.6 & 23.3 & 71.5&44.1&52.7& 72.2\\
    +SFT baseline &  - & 57.9 & 62.4& 20.3& 80.3& 47.5& 58.3 & 78.7 \\
    + Mixed-R1 (Ours)&\cmark &59.2&66.2&23.4 & 78.8&48.6&59.2&79.2 \\
    \midrule
    \multirow{2}{*}{Qwen2.5-VL-7B}& - & 62.4 & 68.9 & 25.9 & 84.1 & 50.6 & 62.3 & 82.7\\
    & \cmark &60.6&64.7 &25.0& 82.1 &50.3 & 60.7& 80.5\\
    +SFT baseline & - & 62.9&68.9&24.3&83.9&52.3&64.7&83.3\\
    +Mixed-R1 (Ours) &\cmark& 64.8& 70.6& 30.3& 84.5&	55.7&64.6 & 83.2 \\
    \bottomrule
    \end{tabular}}
    \caption{Experiment results using Qwen2.5-VL baseline.}
    \label{tab:main_table}
\end{table}

\subsection{Ablation Study and Experiment Analysis}
\label{sec:exp_ablation}

\begin{table}[t]
    \centering
    \adjustbox{max width=0.90\linewidth}{
    \begin{tabular}{l|ccccccccc}
    \toprule
     Data and Reward & Avg. & MathVista & MathVision  & MMMU  & MMStar \\
     \midrule
     Qwen2.5-VL-3B (baseline) & 46.3  & 61.8 & 20.7 & 47.1 & 55.7\\
      +YORN  &44.2 & 57.2& 20.6 & 45.0 & 53.9 \\
      +YORN +MCQ & 47.2 & 61.3 & 21.0 & 49.8 & 56.8 \\ 
      +YORN +MCQ +Chart & 48.0&63.6 & 23.2 & 49.0 & 57.8\\
      +YORN +MCQ +Chart +IoU & 48.0 & 64.3 & 23.3 & 47.6 & 56.8 \\
      +YORN +MCQ +Chart +IoU +Open-Ended & 49.2 & 66.2 & 23.4 & 48.6 & 59.2\\
    \bottomrule
    \end{tabular}}
    \caption{Ablation on the role of each reward. We conduct training by incrementally adding each reward one by one based on an easy-to-hard strategy.}
    \label{tab:rewards}
\end{table}

\begin{table}[t]
\setlength\tabcolsep{2pt}
    \centering
    \vspace{-5mm}
    \begin{subtable}[t]{0.48\textwidth}
        \centering
        \scalebox{0.8}{
        \begin{tabular}{cc|ccc}
            \toprule
            Method & Data Scale & MathVista & MathVision & MMStar\\
            \midrule
            Mixed-R1 & 20K & 62.8 &22.0& 56.1 \\
            Mixed-R1 & 45K &  66.2 & 23.4 & 59.2\\
            Mixed-R1 & 90K & 63.7 & 24.3 & 59.3 \\
            \bottomrule
        \end{tabular}
        }
        \caption{Ablation study on data scale for the Mixed-45K.}
    \end{subtable}
    \hfill
    \begin{subtable}[t]{0.48\textwidth}
        \centering
        \scalebox{0.8}{
        \begin{tabular}{l|ccc}
            \toprule
            Method & MathVista & MathVision &MMStar\\
            \midrule
            InternVL2.5-4B &65.9 & 25.0& 58.7 \\
            +SFT & 51.7& 22.4 & 50.9 \\
            +Mixed R1 (Ours) & 66.2 & 25.1 & 61.9\\
            \bottomrule
        \end{tabular}
        }
        \caption{Ablation study on InternVL2.5-4B.}
    \end{subtable}
    \caption{More ablation and analysis on Mixed-45K and generalization of mixed reward.}
    \label{tab:more_ablation_studies}
\end{table}

\textbf{Ablation on Combination of Different Rewards.}
As our main contribution is the use of multiple rewards for GRPO~\cite{grpo,deepseekr1} post-training, we train our model by incrementally adding rewards one by one based on an easy-to-hard strategy to validate the utility of each reward.
Note that the corresponding data types are also added by each step.
As shown in the Tab.~\ref{tab:rewards}, training with only YORN data results in poor performance.
The main reason is the low data volume, as YORN only has 8K samples. 
This is consistent with the findings in the data scale experiments in Tab.~\ref{tab:more_ablation_studies} (a). 
When more difficult MCQ data is added, the model's performance improves significantly on multiple benchmarks. 
There is an average increase of 3\% on four benchmarks.
Furthermore, as we continue to incrementally add more rewards such as Chart, IoU, and Open-Ended data, the model's performance continues to exhibit upward trends. 
Including IoU refines the model's understanding of spatial relationships and object-level comparisons, which also slightly improves the performance.
Finally, the addition of Open-Ended data pushes the model to handle more unstructured and long-form question-answering scenarios, leading to the highest average score of 49.2. 


\begin{figure}[t]
    \centering
    \includegraphics[width=1.0\linewidth]{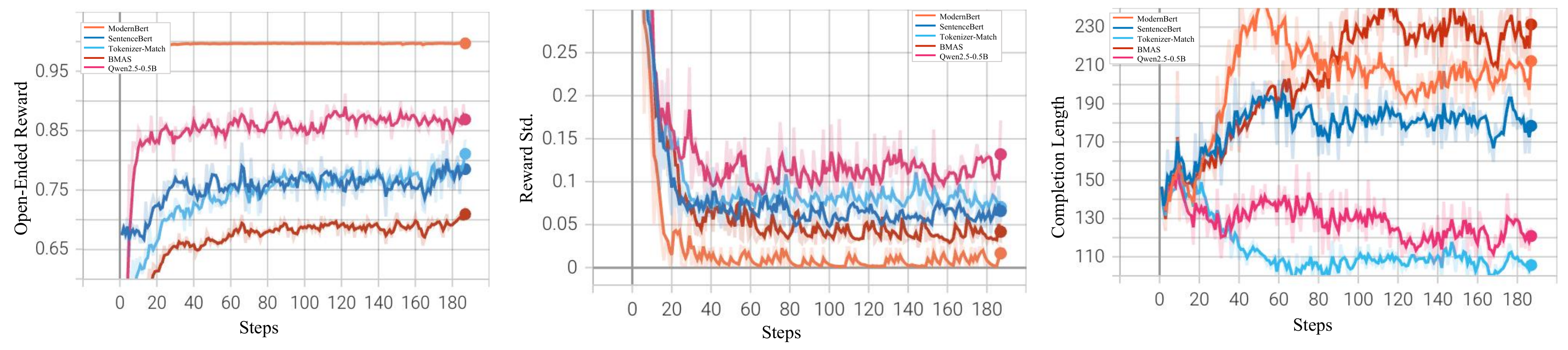}
    \caption{Comparison of different Open-Ended reward designs on Qwen2.5-VL-3B. }
    \label{fig:caption}
\end{figure}

\begin{table}[t]
    \centering
    \begin{tabular}{c|cccccc}
    \toprule
    Open-Ended Reward Design & Avg. & MathVista & MathVision & MMMU & MMStar   \\
    \midrule
    ModernBert & 43.7 & 55.0 & 20.3 & 46.8  & 52.6 \\
    Sentence-Bert & 43.9 & 54.3 & 21.7 & 46.1 &  53.3 \\
    Qwen2.5 (0.5B) & 43.3& 54.8 & 21.4 & 45.1  & 52.0 \\
    Tokenizer bipartite matching & 43.7 & 55.7 & 21.0 & 45.8  & 52.3 \\
    Our BMAS & 44.7 & 57.4 & 20.1 & 47.0  & 54.3 \\
    \bottomrule
    \end{tabular}
    \caption{Ablation on open-ended reward designs. We find that simply using the max scores by tokenizer embeddings can work well.}
    \label{tab:caption}
    \vspace{-3mm}
\end{table}

\noindent
\textbf{Ablation on Open-Ended Reward Design.}
\textit{Comparison setup.} We mainly explore four different setups for open-ended reward design, including both the BERT~\cite{bert,modernbert,sentencebert} and Qwen~\cite{qwen} LLM as a ``judge''.

\noindent
\textbf{\textit{Bert-like model as reward.}}
(1) For the ModernBert~\cite{modernbert} model, we obtain corresponding hidden embeddings $H_{p}^i\in \mathbb{R}^D$ for each token. Subsequently, we calculate an embedding by averaging all token-level hidden embeddings to represent the entire response. Specifically,  $H_p = \frac{1}{n}\sum_i^nH_p^i$, where $n$ denotes the number of tokens in the response. Similarly, we apply the same operation to obtain the $H_{gt}$ for the ground truth.
(2) For the Sentence-BERT~\cite{sentencebert}, it can directly generate a feature representation for each response, eliminating the need for additional averaging operations.  Finally, we compute the cosine similarity between the $H_p$ and $H_{gt}$ and use it as the final reward.

\noindent
\textbf{\textit{Extra LLM as reward.}}
(3) For the Qwen~\cite{qwen} model, we directly feed both the model's response and the ground truth label into the model, prompting it to output a score between 0 and 1 as the reward. The higher the similarity between the response and the ground truth, the higher the score.

\noindent
\textbf{\textit{Use model self-tokenizer distance as reward.}} (4), We further leverage the model's own input embeddings to extract features corresponding to the response. We explore two schemes: bipartite matching and our proposed bidirectional max-average similarity (BMAS) matching. We first compute the similarity between each response token and each ground truth token. 
For the former, we apply bipartite graph matching to find the optimal one-to-one correspondence between tokens. 
For the latter, we calculate bidirectional max-average similarity. 
%

%

\noindent
\textbf{Results and Analysis.} As shown in Fig.~\ref{fig:caption}, we first compare training curves obtained through different approaches. ModernBert consistently maintains a very high reward throughout the training process, indicating that it lacks any discriminative power over different responses. 
Other models show similar trends, with their open-ended rewards steadily increasing over time, among which Qwen2.5-0.5B demonstrates the fastest growth. However, as training progresses, using Qwen2.5-0.5B as the reward function fails to encourage an increase in completion length. 
Overall, only BMAS meets our expectations, and the recovery length gradually increases with training, leading to stable results.
As for the final performance shown in Tab.~\ref{tab:caption}, the different designs of open-ended rewards demonstrate very similar performance across multiple benchmarks. 
Among them, our BMAS achieves the best results, outperforming others by approximately 1\%, which means the most matched words play a key role in building the open-ended rewards. 

\noindent
\textbf{Ablation on Scale of Training Dataset.}
We also explore the performance of Mixed-R1 when trained with different amounts of data in this work. As shown in Tab.~\ref{tab:more_ablation_studies}, we use 20K, 45K, and 90K datasets to train on Qwen2.5VL-3B~\cite{qwen2.5vl}. 
Except for the different data volumes, all other experimental settings remain the same. In the selection of the dataset, we take the 45K data as the baseline, reducing it by half or doubling it for each category of data. 
From the experimental results in Tab.~\ref{tab:more_ablation_studies}, the performance of Mixed-R1 with 20K data has not yet reached saturation.
Using 90K data led to slight improvements in MathVision~\cite{mathvision} and MMStar~\cite{mmstar}, but caused a decline in MathVista~\cite{mathvista}. 
Considering both performance and training costs comprehensively, we finally chose 45K as the training data.

\begin{figure}[t]
    \centering
    \includegraphics[width=1.0\linewidth]{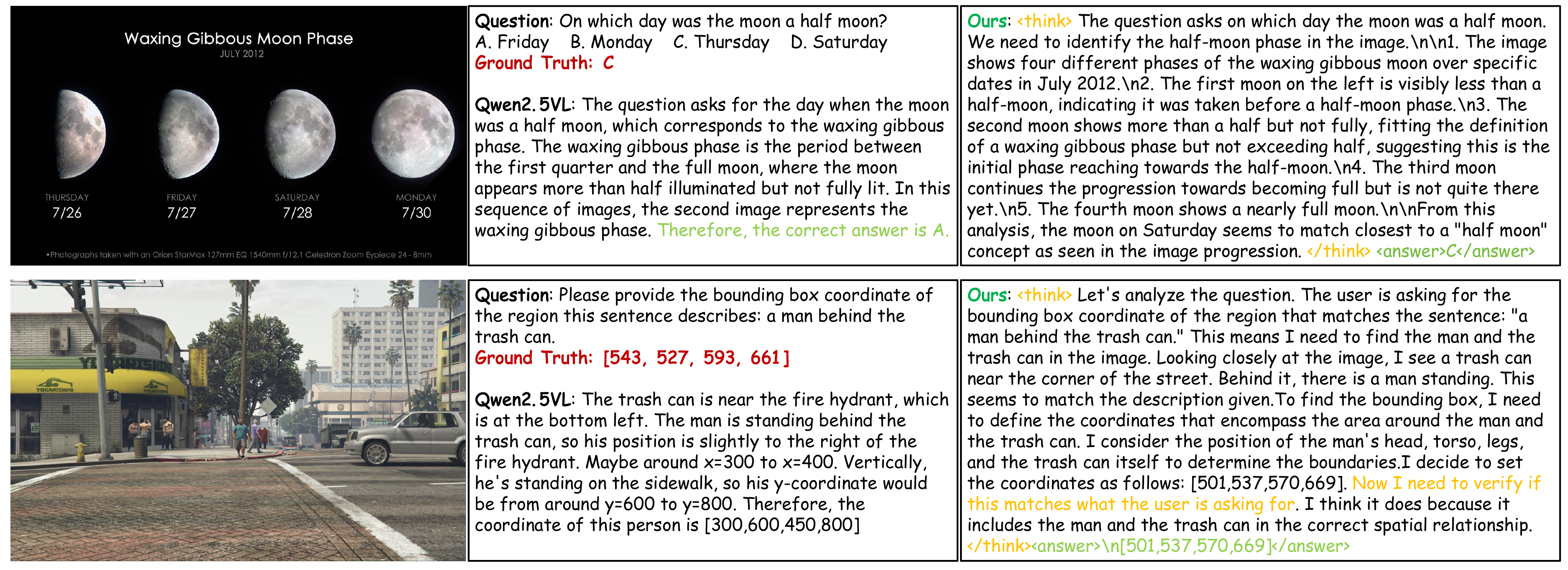}
    \caption{Comparison examples of our Mixed-R1 compared with Qwen2.5-VL-7B-Instruct. We highlight the related reasoning words in orange.}
    \label{fig:example}
\end{figure}

\noindent
\textbf{Visual Qualitative Results.}
Previous studies have found that models trained with rule-based RL methods often exhibit "aha moments," manifesting as behaviors such as reflection and confirmation during prediction.
We also observed a similar phenomenon in Mixed-R1. 
As shown in Fig~\ref{fig:example}, when the model answers questions based on image content, it reconfirms whether the output answer matches the image content and the question before finalizing the response.

\section{Conclusion}
\label{sec:conclusion}

In this work, we present Mixed-R1, the first unified reward design for various SOTA MLLMs.
We collect a high-quality post-training dataset, Mixed-45K, containing five data types, which can be used as a data engine to boost existing MLLMs.
Then, we present a novel tokenizer-driven matching reward to handle the unstable and inferior results caused by the caption data.
Extensive experiments show the effectiveness of our Mixed-R1 framework on various MLLMs.
Ablation studies and analysis indicate the effectiveness of open-ended reward design.
We hope our work can inspire the community to pursue a unified and general design for MLLMs post-training.

\noindent
\textbf{Limitation and future work.} We only conduct experiments on image datasets due to the limited computational resources. 
Future work will extend our unified reward design for image, video, and multiple image inputs.

{\small
\bibliographystyle{plain}
  \bibliography{refbib}

\begin{thebibliography}{10}

\bibitem{gpt4}
Josh Achiam, Steven Adler, Sandhini Agarwal, Lama Ahmad, Ilge Akkaya, Florencia~Leoni Aleman, Diogo Almeida, Janko Altenschmidt, Sam Altman, Shyamal Anadkat, et~al.
\newblock Gpt-4 technical report.
\newblock {\em arXiv preprint arXiv:2303.08774}, 2023.

\bibitem{qwen}
Jinze Bai, Shuai Bai, Yunfei Chu, Zeyu Cui, Kai Dang, Xiaodong Deng, Yang Fan, Wenhang Ge, Yu~Han, Fei Huang, Binyuan Hui, Luo Ji, Mei Li, Junyang Lin, Runji Lin, Dayiheng Liu, Gao Liu, Chengqiang Lu, K.~Lu, Jianxin Ma, Rui Men, Xingzhang Ren, Xuancheng Ren, Chuanqi Tan, Sinan Tan, Jianhong Tu, Peng Wang, Shijie Wang, Wei Wang, Shengguang Wu, Benfeng Xu, Jin Xu, An~Yang, Hao Yang, Jian Yang, Jian Yang, Shusheng Yang, Yang Yao, Bowen Yu, Yu~Bowen, Hongyi Yuan, Zheng Yuan, Jianwei Zhang, Xing Zhang, Yichang Zhang, Zhenru Zhang, Chang Zhou, Jingren Zhou, Xiaohuan Zhou, and Tianhang Zhu.
\newblock Qwen technical report.
\newblock {\em arXiv preprint arXiv:2309.16609}, 2023.

\bibitem{qwen2.5vl}
Shuai Bai, Keqin Chen, Xuejing Liu, Jialin Wang, Wenbin Ge, Sibo Song, Kai Dang, Peng Wang, Shijie Wang, Jun Tang, et~al.
\newblock Qwen2. 5-vl technical report.
\newblock {\em arXiv preprint arXiv:2502.13923}, 2025.

\bibitem{mapqa}
Shuaichen Chang, David Palzer, Jialin Li, Eric Fosler-Lussier, and Ningchuan Xiao.
\newblock Mapqa: A dataset for question answering on choropleth maps.
\newblock {\em arXiv preprint arXiv:2211.08545}, 2022.

\bibitem{allava}
Guiming~Hardy Chen, Shunian Chen, Ruifei Zhang, Junying Chen, Xiangbo Wu, Zhiyi Zhang, Zhihong Chen, Jianquan Li, Xiang Wan, and Benyou Wang.
\newblock Allava: Harnessing gpt4v-synthesized data for lite vision-language models.
\newblock {\em arXiv preprint arXiv:2402.11684}, 2024.

\bibitem{unigeo}
Jiaqi Chen, Tong Li, Jinghui Qin, Pan Lu, Liang Lin, Chongyu Chen, and Xiaodan Liang.
\newblock Unigeo: Unifying geometry logical reasoning via reformulating mathematical expression.
\newblock {\em arXiv preprint arXiv:2212.02746}, 2022.

\bibitem{geoqa}
Jiaqi Chen, Jianheng Tang, Jinghui Qin, Xiaodan Liang, Lingbo Liu, Eric~P Xing, and Liang Lin.
\newblock Geoqa: A geometric question answering benchmark towards multimodal numerical reasoning.
\newblock {\em arXiv preprint arXiv:2105.14517}, 2021.

\bibitem{r1v}
Liang Chen, Lei Li, Haozhe Zhao, Yifan Song, and Vinci.
\newblock R1-v: Reinforcing super generalization ability in vision-language models with less than \$3.
\newblock \url{https://github.com/Deep-Agent/R1-V}, 2025.

\bibitem{mmstar}
Lin Chen, Jinsong Li, Xiaoyi Dong, Pan Zhang, Yuhang Zang, Zehui Chen, Haodong Duan, Jiaqi Wang, Yu~Qiao, Dahua Lin, et~al.
\newblock Are we on the right way for evaluating large vision-language models?
\newblock {\em arXiv preprint arXiv:2403.20330}, 2024.

\bibitem{intern2.5}
Zhe Chen, Weiyun Wang, Yue Cao, Yangzhou Liu, Zhangwei Gao, Erfei Cui, Jinguo Zhu, Shenglong Ye, Hao Tian, Zhaoyang Liu, et~al.
\newblock Expanding performance boundaries of open-source multimodal models with model, data, and test-time scaling.
\newblock {\em arXiv preprint arXiv:2412.05271}, 2024.

\bibitem{vicuna}
Wei-Lin Chiang, Zhuohan Li, Zi~Lin, Ying Sheng, Zhanghao Wu, Hao Zhang, Lianmin Zheng, Siyuan Zhuang, Yonghao Zhuang, Joseph~E. Gonzalez, Ion Stoica, and Eric~P. Xing.
\newblock Vicuna: An open-source chatbot impressing gpt-4 with 90\%* chatgpt quality, 2023.

\bibitem{deepseekr1}
DeepSeek-AI.
\newblock Deepseek-r1: Incentivizing reasoning capability in llms via reinforcement learning, 2025.

\bibitem{navit}
Mostafa Dehghani, Basil Mustafa, Josip Djolonga, Jonathan Heek, Matthias Minderer, Mathilde Caron, Andreas Steiner, Joan Puigcerver, Robert Geirhos, Ibrahim~M Alabdulmohsin, et~al.
\newblock Patch n’pack: Navit, a vision transformer for any aspect ratio and resolution.
\newblock In {\em NeurIPS}, 2023.

\bibitem{bert}
Jacob Devlin, Ming-Wei Chang, Kenton Lee, and Kristina Toutanova.
\newblock Bert: Pre-training of deep bidirectional transformers for language understanding.
\newblock In {\em NAACL}, 2019.

\bibitem{fei2025path}
Hao Fei, Yuan Zhou, Juncheng Li, Xiangtai Li, Qingshan Xu, Bobo Li, Shengqiong Wu, Yaoting Wang, Junbao Zhou, Jiahao Meng, et~al.
\newblock On path to multimodal generalist: General-level and general-bench.
\newblock {\em arXiv preprint arXiv:2505.04620}, 2025.

\bibitem{videomme}
Chaoyou Fu, Yuhan Dai, Yongdong Luo, Lei Li, Shuhuai Ren, Renrui Zhang, Zihan Wang, Chenyu Zhou, Yunhang Shen, Mengdan Zhang, et~al.
\newblock Video-mme: The first-ever comprehensive evaluation benchmark of multi-modal llms in video analysis.
\newblock {\em arXiv preprint arXiv:2405.21075}, 2024.

\bibitem{vita}
Chaoyou Fu, Haojia Lin, Xiong Wang, Yi-Fan Zhang, Yunhang Shen, Xiaoyu Liu, Haoyu Cao, Zuwei Long, Heting Gao, Ke~Li, et~al.
\newblock Vita-1.5: Towards gpt-4o level real-time vision and speech interaction.
\newblock {\em arXiv preprint arXiv:2501.01957}, 2025.

\bibitem{cantor}
Timin Gao, Peixian Chen, Mengdan Zhang, Chaoyou Fu, Yunhang Shen, Yan Zhang, Shengchuan Zhang, Xiawu Zheng, Xing Sun, Liujuan Cao, and Rongrong Ji.
\newblock Cantor: Inspiring multimodal chain-of-thought of mllm.
\newblock {\em ACM MM}, 2024.

\bibitem{hallusionbench}
Tianrui Guan, Fuxiao Liu, Xiyang Wu, Ruiqi Xian, Zongxia Li, Xiaoyu Liu, Xijun Wang, Lichang Chen, Furong Huang, Yaser Yacoob, et~al.
\newblock Hallusionbench: an advanced diagnostic suite for entangled language hallucination and visual illusion in large vision-language models.
\newblock In {\em CVPR}, 2024.

\bibitem{r1}
Daya Guo, Dejian Yang, Haowei Zhang, Junxiao Song, Ruoyu Zhang, Runxin Xu, Qihao Zhu, Shirong Ma, Peiyi Wang, Xiao Bi, et~al.
\newblock Deepseek-r1: Incentivizing reasoning capability in llms via reinforcement learning.
\newblock {\em arXiv preprint arXiv:2501.12948}, 2025.

\bibitem{han2024free}
Kai Han, Jianyuan Guo, Yehui Tang, Wei He, Enhua Wu, and Yunhe Wang.
\newblock Free video-llm: Prompt-guided visual perception for efficient training-free video llms.
\newblock {\em arXiv preprint arXiv:2410.10441}, 2024.

\bibitem{gpt4o}
Aaron Hurst, Adam Lerer, Adam~P Goucher, Adam Perelman, Aditya Ramesh, Aidan Clark, AJ~Ostrow, Akila Welihinda, Alan Hayes, Alec Radford, et~al.
\newblock Gpt-4o system card.
\newblock {\em arXiv preprint arXiv:2410.21276}, 2024.

\bibitem{Jie2024MemorySpaceVP}
Shibo Jie, Yehui Tang, Ning Ding, Zhi-Hong Deng, Kai Han, and Yunhe Wang.
\newblock Memory-space visual prompting for efficient vision-language fine-tuning.
\newblock {\em arXiv preprint arXiv:2405.05615}, 2025.

\bibitem{clevr}
Justin Johnson, Bharath Hariharan, Laurens Van Der~Maaten, Li~Fei-Fei, C~Lawrence~Zitnick, and Ross Girshick.
\newblock Clevr: A diagnostic dataset for compositional language and elementary visual reasoning.
\newblock In {\em CVPR}, 2017.

\bibitem{figureqa}
Samira~Ebrahimi Kahou, Vincent Michalski, Adam Atkinson, {\'A}kos K{\'a}d{\'a}r, Adam Trischler, and Yoshua Bengio.
\newblock Figureqa: An annotated figure dataset for visual reasoning.
\newblock {\em arXiv preprint arXiv:1710.07300}, 2017.

\bibitem{ai2d}
Aniruddha Kembhavi, Mike Salvato, Eric Kolve, Minjoon Seo, Hannaneh Hajishirzi, and Ali Farhadi.
\newblock A diagram is worth a dozen images.
\newblock In {\em ECCV}, 2016.

\bibitem{llava-ov}
Bo~Li, Yuanhan Zhang, Dong Guo, Renrui Zhang, Feng Li, Hao Zhang, Kaichen Zhang, Peiyuan Zhang, Yanwei Li, Ziwei Liu, et~al.
\newblock Llava-onevision: Easy visual task transfer.
\newblock {\em arXiv preprint arXiv:2408.03326}, 2024.

\bibitem{seedbench}
Bohao Li, Rui Wang, Guangzhi Wang, Yuying Ge, Yixiao Ge, and Ying Shan.
\newblock Seed-bench: Benchmarking multimodal llms with generative comprehension.
\newblock {\em arXiv preprint arXiv:2307.16125}, 2023.

\bibitem{mvbench}
Kunchang Li, Yali Wang, Yinan He, Yizhuo Li, Yi~Wang, Yi~Liu, Zun Wang, Jilan Xu, Guo Chen, Ping Luo, et~al.
\newblock Mvbench: A comprehensive multi-modal video understanding benchmark.
\newblock In {\em CVPR}, 2024.

\bibitem{videochart-r1}
Xinhao Li, Ziang Yan, Desen Meng, Lu~Dong, Xiangyu Zeng, Yinan He, Yali Wang, Yu~Qiao, Yi~Wang, and Limin Wang.
\newblock Videochat-r1: Enhancing spatio-temporal perception via reinforcement fine-tuning.
\newblock {\em arXiv preprint arXiv:2504.06958}, 2025.

\bibitem{deepseekv3}
Aixin Liu, Bei Feng, Bing Xue, Bingxuan Wang, Bochao Wu, Chengda Lu, Chenggang Zhao, Chengqi Deng, Chenyu Zhang, Chong Ruan, et~al.
\newblock Deepseek-v3 technical report.
\newblock {\em arXiv preprint arXiv:2412.19437}, 2024.

\bibitem{vsr}
Fangyu Liu, Guy Emerson, and Nigel Collier.
\newblock Visual spatial reasoning.
\newblock {\em TACL}, 2023.

\bibitem{llava1.5}
Haotian Liu, Chunyuan Li, Yuheng Li, and Yong~Jae Lee.
\newblock Improved baselines with visual instruction tuning.
\newblock In {\em CVPR}, 2024.

\bibitem{llava}
Haotian Liu, Chunyuan Li, Qingyang Wu, and Yong~Jae Lee.
\newblock Visual instruction tuning.
\newblock In {\em NeurIPS}, 2023.

\bibitem{mmbench}
Yuan Liu, Haodong Duan, Yuanhan Zhang, Bo~Li, Songyang Zhang, Wangbo Zhao, Yike Yuan, Jiaqi Wang, Conghui He, Ziwei Liu, et~al.
\newblock Mmbench: Is your multi-modal model an all-around player?
\newblock In {\em ECCV}, 2024.

\bibitem{ocrbench}
Yuliang Liu, Zhang Li, Mingxin Huang, Biao Yang, Wenwen Yu, Chunyuan Li, Xu-Cheng Yin, Cheng-Lin Liu, Lianwen Jin, and Xiang Bai.
\newblock Ocrbench: on the hidden mystery of ocr in large multimodal models.
\newblock {\em SCIS}, 2024.

\bibitem{deepseek-vl}
Haoyu Lu, Wen Liu, Bo~Zhang, Bingxuan Wang, Kai Dong, Bo~Liu, Jingxiang Sun, Tongzheng Ren, Zhuoshu Li, Hao Yang, et~al.
\newblock Deepseek-vl: towards real-world vision-language understanding.
\newblock {\em arXiv preprint arXiv:2403.05525}, 2024.

\bibitem{mathvista}
Pan Lu, Hritik Bansal, Tony Xia, Jiacheng Liu, Chunyuan Li, Hannaneh Hajishirzi, Hao Cheng, Kai-Wei Chang, Michel Galley, and Jianfeng Gao.
\newblock Mathvista: Evaluating mathematical reasoning of foundation models in visual contexts.
\newblock In {\em ICLR}, 2024.

\bibitem{scienceqa}
Pan Lu, Swaroop Mishra, Tony Xia, Liang Qiu, Kai-Wei Chang, Song-Chun Zhu, Oyvind Tafjord, Peter Clark, and Ashwin Kalyan.
\newblock Learn to explain: Multimodal reasoning via thought chains for science question answering.
\newblock In {\em NeurIPS}, 2022.

\bibitem{luo2023cheap}
Gen Luo, Yiyi Zhou, Tianhe Ren, Shengxin Chen, Xiaoshuai Sun, and Rongrong Ji.
\newblock Cheap and quick: Efficient vision-language instruction tuning for large language models.
\newblock {\em NeurIPS}, 2023.

\bibitem{luo2024feast}
Gen Luo, Yiyi Zhou, Yuxin Zhang, Xiawu Zheng, Xiaoshuai Sun, and Rongrong Ji.
\newblock Feast your eyes: Mixture-of-resolution adaptation for multimodal large language models.
\newblock {\em arXiv preprint arXiv:2403.03003}, 2024.

\bibitem{videorag}
Yongdong Luo, Xiawu Zheng, Xiao Yang, Guilin Li, Haojia Lin, Jinfa Huang, Jiayi Ji, Fei Chao, Jiebo Luo, and Rongrong Ji.
\newblock Video-rag: Visually-aligned retrieval-augmented long video comprehension.
\newblock {\em arXiv preprint arXiv:2411.13093}, 2024.

\bibitem{mllm-selector}
Yiwei Ma, Guohai Xu, Xiaoshuai Sun, Jiayi Ji, Jie Lou, Debing Zhang, and Rongrong Ji.
\newblock Mllm-selector: Necessity and diversity-driven high-value data selection for enhanced visual instruction tuning.
\newblock {\em arXiv preprint arXiv:2503.20502}, 2025.

\bibitem{refcocog}
Junhua Mao, Jonathan Huang, Alexander Toshev, Oana Camburu, Alan~L Yuille, and Kevin Murphy.
\newblock Generation and comprehension of unambiguous object descriptions.
\newblock In {\em CVPR}, 2016.

\bibitem{chartqa}
Ahmed Masry, Do~Xuan Long, Jia~Qing Tan, Shafiq Joty, and Enamul Hoque.
\newblock Chartqa: A benchmark for question answering about charts with visual and logical reasoning.
\newblock {\em arXiv preprint arXiv:2203.10244}, 2022.

\bibitem{infoqa}
Minesh Mathew, Viraj Bagal, Rub{\`e}n Tito, Dimosthenis Karatzas, Ernest Valveny, and CV~Jawahar.
\newblock Infographicvqa.
\newblock In {\em WACV}, 2022.

\bibitem{docvqa}
Minesh Mathew, Dimosthenis Karatzas, and CV~Jawahar.
\newblock Docvqa: A dataset for vqa on document images.
\newblock In {\em WACV}, 2021.

\bibitem{mm-eureka}
Fanqing Meng, Lingxiao Du, Zongkai Liu, Zhixiang Zhou, Quanfeng Lu, Daocheng Fu, Tiancheng Han, Botian Shi, Wenhai Wang, Junjun He, et~al.
\newblock Mm-eureka: Exploring the frontiers of multimodal reasoning with rule-based reinforcement learning.
\newblock {\em arXiv preprint arXiv:2503.07365}, 2025.

\bibitem{rlhf}
Long Ouyang, Jeffrey Wu, Xu~Jiang, Diogo Almeida, Carroll Wainwright, Pamela Mishkin, Chong Zhang, Sandhini Agarwal, Katarina Slama, Alex Ray, et~al.
\newblock Training language models to follow instructions with human feedback.
\newblock {\em NeurIPS}, 2022.

\bibitem{clip}
Alec Radford, Jong~Wook Kim, Chris Hallacy, Aditya Ramesh, Gabriel Goh, Sandhini Agarwal, Girish Sastry, Amanda Askell, Pamela Mishkin, Jack Clark, et~al.
\newblock Learning transferable visual models from natural language supervision.
\newblock In {\em ICML}, 2021.

\bibitem{dpo}
Rafael Rafailov, Archit Sharma, Eric Mitchell, Christopher~D Manning, Stefano Ermon, and Chelsea Finn.
\newblock Direct preference optimization: Your language model is secretly a reward model.
\newblock {\em NeurIPS}, 2023.

\bibitem{evem}
Miao Rang, Zhenni Bi, Chuanjian Liu, Yehui Tang, Kai Han, and Yunhe Wang.
\newblock Eve: Efficient multimodal vision language models with elastic visual experts.
\newblock {\em arXiv preprint arXiv:2501.04322}, 2025.

\bibitem{sentencebert}
Nils Reimers and Iryna Gurevych.
\newblock Sentence-bert: Sentence embeddings using siamese bert-networks.
\newblock {\em arXiv preprint arXiv:1908.10084}, 2019.

\bibitem{ppo}
John Schulman, Filip Wolski, Prafulla Dhariwal, Alec Radford, and Oleg Klimov.
\newblock Proximal policy optimization algorithms.
\newblock {\em arXiv preprint arXiv:1707.06347}, 2017.

\bibitem{grpo}
Zhihong Shao, Peiyi Wang, Qihao Zhu, Runxin Xu, Junxiao Song, Xiao Bi, Haowei Zhang, Mingchuan Zhang, YK~Li, Y~Wu, et~al.
\newblock Deepseekmath: Pushing the limits of mathematical reasoning in open language models.
\newblock {\em arXiv preprint arXiv:2402.03300}, 2024.

\bibitem{vlm-r1}
Haozhan Shen, Peng Liu, Jingcheng Li, Chunxin Fang, Yibo Ma, Jiajia Liao, Qiaoli Shen, Zilun Zhang, Kangjia Zhao, Qianqian Zhang, Ruochen Xu, and Tiancheng Zhao.
\newblock Vlm-r1: A stable and generalizable r1-style large vision-language model.
\newblock {\em arXiv preprint arXiv:2504.07615}, 2025.

\bibitem{long-vita}
Yunhang Shen, Chaoyou Fu, Shaoqi Dong, Xiong Wang, Peixian Chen, Mengdan Zhang, Haoyu Cao, Ke~Li, Xiawu Zheng, Yan Zhang, et~al.
\newblock Long-vita: Scaling large multi-modal models to 1 million tokens with leading short-context accuray.
\newblock {\em arXiv preprint arXiv:2502.05177}, 2025.

\bibitem{gemini}
Gemini Team, Rohan Anil, Sebastian Borgeaud, Jean-Baptiste Alayrac, Jiahui Yu, Radu Soricut, Johan Schalkwyk, Andrew~M Dai, Anja Hauth, Katie Millican, et~al.
\newblock Gemini: a family of highly capable multimodal models.
\newblock {\em arXiv preprint arXiv:2312.11805}, 2023.

\bibitem{llama}
Hugo Touvron, Thibaut Lavril, Gautier Izacard, Xavier Martinet, Marie-Anne Lachaux, Timoth{\'e}e Lacroix, Baptiste Rozi{\`e}re, Naman Goyal, Eric Hambro, Faisal Azhar, et~al.
\newblock Llama: Open and efficient foundation language models.
\newblock {\em arXiv preprint arXiv:2302.13971}, 2023.

\bibitem{mathvision}
Ke~Wang, Junting Pan, Weikang Shi, Zimu Lu, Houxing Ren, Aojun Zhou, Mingjie Zhan, and Hongsheng Li.
\newblock Measuring multimodal mathematical reasoning with math-vision dataset.
\newblock {\em arXiv preprint arXiv:2402.14804}, 2024.

\bibitem{qwenvl}
Peng Wang, Shuai Bai, Sinan Tan, Shijie Wang, Zhihao Fan, Jinze Bai, Keqin Chen, Xuejing Liu, Jialin Wang, Wenbin Ge, et~al.
\newblock Qwen2-vl: Enhancing vision-language model's perception of the world at any resolution.
\newblock {\em arXiv preprint arXiv:2409.12191}, 2024.

\bibitem{modernbert}
Benjamin Warner, Antoine Chaffin, Benjamin Clavi{\'e}, Orion Weller, Oskar Hallstr{\"o}m, Said Taghadouini, Alexis Gallagher, Raja Biswas, Faisal Ladhak, Tom Aarsen, et~al.
\newblock Smarter, better, faster, longer: A modern bidirectional encoder for fast, memory efficient, and long context finetuning and inference.
\newblock {\em arXiv preprint arXiv:2412.13663}, 2024.

\bibitem{qwen2.5-1m}
An~Yang, Bowen Yu, Chengyuan Li, Dayiheng Liu, Fei Huang, Haoyan Huang, Jiandong Jiang, Jianhong Tu, Jianwei Zhang, Jingren Zhou, Junyang Lin, Kai Dang, Kexin Yang, Le~Yu, Mei Li, Minmin Sun, Qin Zhu, Rui Men, Tao He, Weijia Xu, Wenbiao Yin, Wenyuan Yu, Xiafei Qiu, Xingzhang Ren, Xinlong Yang, Yong Li, Zhiying Xu, and Zipeng Zhang.
\newblock Qwen2.5-1m technical report.
\newblock {\em arXiv preprint arXiv:2501.15383}, 2025.

\bibitem{r1-onevision}
Yi~Yang, Xiaoxuan He, Hongkun Pan, Xiyan Jiang, Yan Deng, Xingtao Yang, Haoyu Lu, Dacheng Yin, Fengyun Rao, Minfeng Zhu, et~al.
\newblock R1-onevision: Advancing generalized multimodal reasoning through cross-modal formalization.
\newblock {\em arXiv preprint arXiv:2503.10615}, 2025.

\bibitem{minicpm}
Yuan Yao, Tianyu Yu, Ao~Zhang, Chongyi Wang, Junbo Cui, Hongji Zhu, Tianchi Cai, Haoyu Li, Weilin Zhao, Zhihui He, et~al.
\newblock Minicpm-v: A gpt-4v level mllm on your phone.
\newblock {\em arXiv preprint arXiv:2408.01800}, 2024.

\bibitem{mmvet}
Weihao Yu, Zhengyuan Yang, Linjie Li, Jianfeng Wang, Kevin Lin, Zicheng Liu, Xinchao Wang, and Lijuan Wang.
\newblock Mm-vet: Evaluating large multimodal models for integrated capabilities.
\newblock {\em arXiv preprint arXiv:2308.02490}, 2023.

\bibitem{yuan2025sa2va}
Haobo Yuan, Xiangtai Li, Tao Zhang, Zilong Huang, Shilin Xu, Shunping Ji, Yunhai Tong, Lu~Qi, Jiashi Feng, and Ming-Hsuan Yang.
\newblock Sa2va: Marrying sam2 with llava for dense grounded understanding of images and videos.
\newblock {\em arXiv preprint arXiv:2501.04001}, 2025.

\bibitem{mmmu}
Xiang Yue, Yuansheng Ni, Kai Zhang, Tianyu Zheng, Ruoqi Liu, Ge~Zhang, Samuel Stevens, Dongfu Jiang, Weiming Ren, Yuxuan Sun, Cong Wei, Botao Yu, Ruibin Yuan, Renliang Sun, Ming Yin, Boyuan Zheng, Zhenzhu Yang, Yibo Liu, Wenhao Huang, Huan Sun, Yu~Su, and Wenhu Chen.
\newblock Mmmu: A massive multi-discipline multimodal understanding and reasoning benchmark for expert agi.
\newblock In {\em CVPR}, 2024.

\bibitem{siglip}
Xiaohua Zhai, Basil Mustafa, Alexander Kolesnikov, and Lucas Beyer.
\newblock Sigmoid loss for language image pre-training.
\newblock In {\em ICCV}, 2023.

\bibitem{vision-r1}
Yufei Zhan, Yousong Zhu, Shurong Zheng, Hongyin Zhao, Fan Yang, Ming Tang, and Jinqiao Wang.
\newblock Vision-r1: Evolving human-free alignment in large vision-language models via vision-guided reinforcement learning.
\newblock {\em arXiv preprint arXiv:2503.18013}, 2025.

\bibitem{lmms-eval}
Kaichen Zhang, Bo~Li, Peiyuan Zhang, Fanyi Pu, Joshua~Adrian Cahyono, Kairui Hu, Shuai Liu, Yuanhan Zhang, Jingkang Yang, Chunyuan Li, and Ziwei Liu.
\newblock Lmms-eval: Reality check on the evaluation of large multimodal models.
\newblock {\em arXiv preprint arXiv:2407.12772}, 2024.

\bibitem{zhang2024omg}
Tao Zhang, Xiangtai Li, Hao Fei, Haobo Yuan, Shengqiong Wu, Shunping Ji, Chen~Change Loy, and Shuicheng Yan.
\newblock Omg-llava: Bridging image-level, object-level, pixel-level reasoning and understanding.
\newblock In {\em NeurIPS}, 2024.

\bibitem{zhang2025pixel}
Tao Zhang, Xiangtai Li, Zilong Huang, Yanwei Li, Weixian Lei, Xueqing Deng, Shihao Chen, Shunping Ji, and Jiashi Feng.
\newblock Pixel-sail: Single transformer for pixel-grounded understanding.
\newblock {\em arXiv preprint arXiv:2504.10465}, 2025.

\bibitem{zhang2024enhancing}
Xiaofeng Zhang, Fanshuo Zeng, Yihao Quan, Zheng Hui, and Jiawei Yao.
\newblock Enhancing multimodal large language models complex reason via similarity computation.
\newblock {\em AAAI}, 2024.

\bibitem{multihiertt}
Yilun Zhao, Yunxiang Li, Chenying Li, and Rui Zhang.
\newblock {M}ulti{H}iertt: Numerical reasoning over multi hierarchical tabular and textual data.
\newblock In {\em ACL}, 2022.

\bibitem{mluv}
Junjie Zhou, Yan Shu, Bo~Zhao, Boya Wu, Shitao Xiao, Xi~Yang, Yongping Xiong, Bo~Zhang, Tiejun Huang, and Zheng Liu.
\newblock Mlvu: A comprehensive benchmark for multi-task long video understanding.
\newblock {\em arXiv preprint arXiv:2406.04264}, 2024.

\bibitem{zhou2025they}
Yikang Zhou, Tao Zhang, Shilin Xu, Shihao Chen, Qianyu Zhou, Yunhai Tong, Shunping Ji, Jiangning Zhang, Xiangtai Li, and Lu~Qi.
\newblock Are they the same? exploring visual correspondence shortcomings of multimodal llms.
\newblock {\em arXiv preprint arXiv:2501.04670}, 2025.

\bibitem{internvl3}
Jinguo Zhu, Weiyun Wang, Zhe Chen, Zhaoyang Liu, Shenglong Ye, Lixin Gu, Yuchen Duan, Hao Tian, Weijie Su, Jie Shao, et~al.
\newblock Internvl3: Exploring advanced training and test-time recipes for open-source multimodal models.
\newblock {\em arXiv preprint arXiv:2504.10479}, 2025.

\bibitem{zhu2023genimage}
Mingjian Zhu, Hanting Chen, Qiangyu Yan, Xudong Huang, Guanyu Lin, Wei Li, Zhijun Tu, Hailin Hu, Jie Hu, and Yunhe Wang.
\newblock Genimage: A million-scale benchmark for detecting ai-generated image.
\newblock {\em NeurIPS}, 2023.

\end{thebibliography}
}

\end{document}